\documentclass[conference]{IEEEtran}
\IEEEoverridecommandlockouts
\usepackage[noadjust]{cite}

\usepackage{amsmath,amssymb,amsfonts}
\usepackage{graphicx}
\usepackage{textcomp}
\usepackage{xcolor}
\usepackage{caption}
\usepackage{algorithm}
\usepackage{algpseudocode}
\usepackage{booktabs}
\usepackage{bm}
\usepackage{hyperref}
\usepackage{balance}
\def\BibTeX{{\rm B\kern-.05em{\sc i\kern-.025em b}\kern-.08em
    T\kern-.1667em\lower.7ex\hbox{E}\kern-.125emX}}
\newcommand{\rv}[1]{{#1}}
\newcommand{\blanklines}[1]{%
  \par\noindent
  \loop\ifnum\value{line}<#1
    \\ 
    \addtocounter{line}{1} 
  \repeat
}
\newcounter{line}

\begin{document}

\title{\textbf{GRACE}: \textbf{G}eneralizing \textbf{R}obot-\textbf{A}ssisted \textbf{C}aregiving with User Functionality \textbf{E}mbeddings\\
}

\author{\IEEEauthorblockN{Ziang Liu\IEEEauthorrefmark{2}, Yuanchen Ju\IEEEauthorrefmark{2}, Yu Da\IEEEauthorrefmark{2}, Tom Silver\IEEEauthorrefmark{2}, Pranav N. Thakkar\IEEEauthorrefmark{2}, Jenna Li\IEEEauthorrefmark{2}, Justin Guo\IEEEauthorrefmark{2}\\ Katherine Dimitropoulou\IEEEauthorrefmark{4}, Tapomayukh Bhattacharjee\IEEEauthorrefmark{2}} \IEEEauthorblockA{\IEEEauthorrefmark{2}\textit{Cornell University}, Ithaca, NY, USA\\
\{zl873, yj567, yd373, tss95, pnt8, yl3647, jjg283, tapomayukh\}@cornell.edu} \IEEEauthorblockA{\IEEEauthorrefmark{4}\textit{Columbia University}, New York City, NY, USA\\
kd2524@cumc.columbia.edu}
}

\twocolumn[{%
\renewcommand\twocolumn[1][]{#1}%

\maketitle
\thispagestyle{empty}
\pagestyle{empty}
\begin{center}
    \begin{minipage}{1.0\textwidth}
        \centering
        \vspace{-0.5cm}
        \includegraphics[width=\textwidth, trim=12mm 35mm 22mm 16mm, clip]{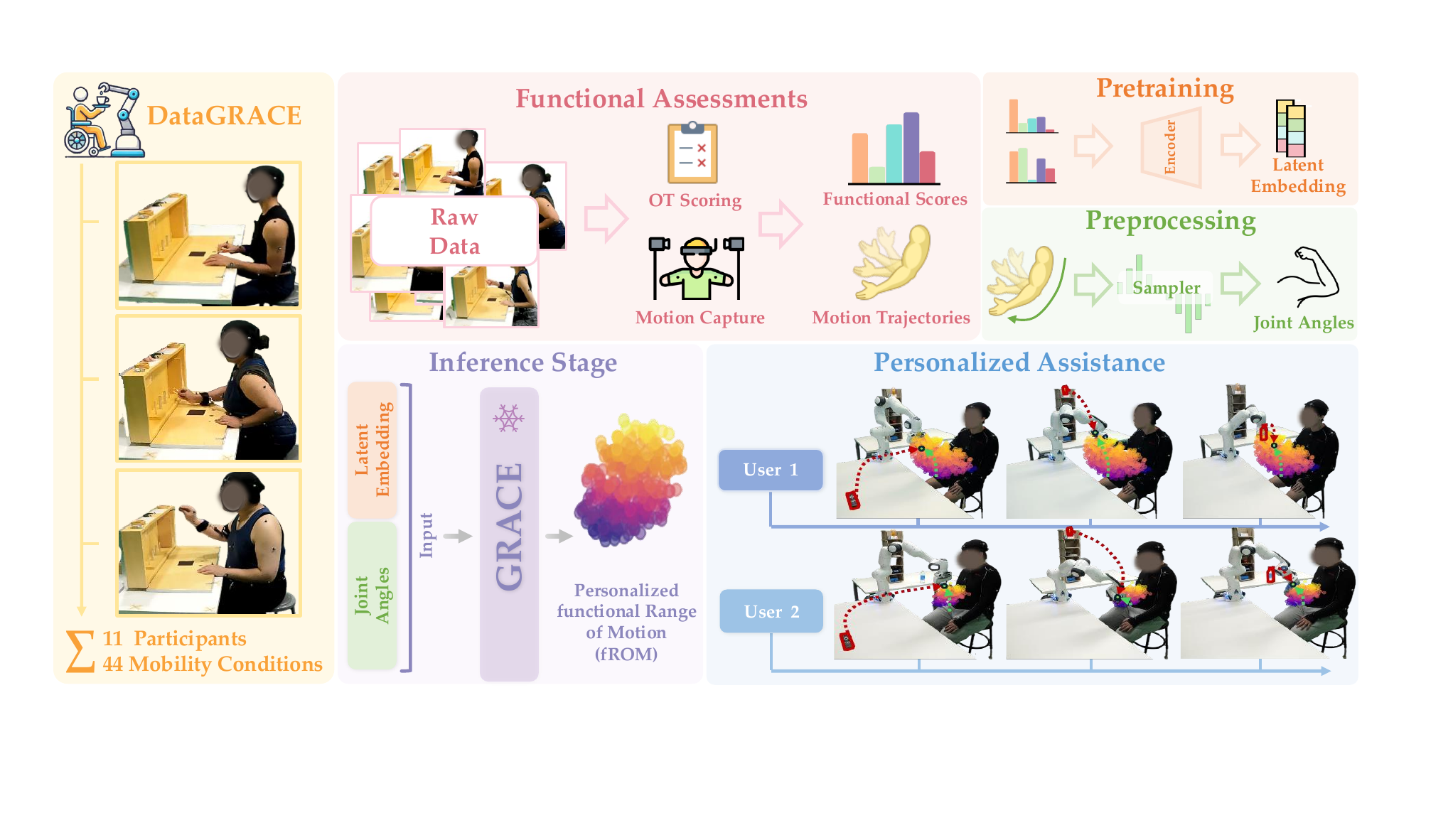}
        \vspace{-0.7cm}
        \captionof{figure}{\small We present GRACE, a method for generalizing robot-assisted caregiving by creating personalized user models through predicting functional range of motion (fROM) from functional assessments.}
        \label{fig:teaser}
        \vspace{-0.2cm}
    \end{minipage}
\end{center}
}]

\footnotetext[1]{This work was partly funded by NSF IIS \#2132846 and CAREER \#2238792. We thank Rajat Jenamani for their help with problem formulation and visualizations, and Jovan Menezes for their help with data collection setup.}

\begin{abstract}
Robot caregiving should be personalized to meet the diverse needs of care recipients---assisting with tasks as needed, while taking user agency in action into account.
In physical tasks such as handover, bathing, dressing, and rehabilitation, a key aspect of this diversity is the functional range of motion (fROM), which can vary significantly between individuals.
In this work, we learn to predict personalized fROM as a way to generalize robot decision-making in a wide range of caregiving tasks.
We propose a novel data-driven method for predicting personalized fROM using functional assessment scores from occupational therapy.
We develop a neural model that learns to embed functional assessment scores into a latent representation of the user's physical function.
The model is trained using motion capture data collected from users with emulated mobility limitations.
After training, the model predicts personalized fROM for new users without motion capture.
Through simulated experiments and a real-robot user study, we show that the personalized fROM predictions from our model enable the robot to provide personalized and effective assistance while improving the user's agency in action. See our website for more visualizations: \url{https://emprise.cs.cornell.edu/grace/}.

\end{abstract}


\newcommand{\user}{u}
\newcommand{\functionalscores}{\boldsymbol{\sigma}}
\newcommand{\latent}{\mathbf{z}}
\newcommand{\jointstate}{\bm{\theta}}
\newcommand{\from}{\Theta}
\newcommand{\classificationinput}{\mathbf{x}}
\newcommand{\classificationoutput}{y}
\newcommand{\classificationdataset}{\mathcal{D}}
\newcommand{\datasetname}{DataGRACE}

\vspace{-0.3cm}
\section{Introduction}

\vspace{-0.1cm}

Participants with Spinal Muscular Atrophy gave examples that \emph{``stressed the importance of personalized adaptations. These insights highlight the need for adaptive algorithms and interfaces that are tailored to individual needs, moving away from a one-size-fits-all model.''}~\cite{jenamani2024feel}

Care recipients and caregivers alike voice the need for \emph{personalization} in assistive robotics~\cite{bedaf2018multi,madan2022sparcs,nanavati2023physically}.
If the goal is to empower individuals, then ``one size does not fit all.'' 
Recent advances in caregiving robots offer specific solutions for \emph{individual users} in activities of daily living (ADLs) such as feeding~\cite{feeding1,feeding2,feeding3}, dressing~\cite{dressing1,dressing2}, bathing~\cite{bathing1,bathing2}, transferring~\cite{transferring1}, and ambulating~\cite{ambulation1}.
To create caregiving robots at the scale necessary to meet the needs of the approximately 1.3 billion people worldwide who live with significant disabilities~\cite{who2022}, we should develop fundamental methods that span multiple activities and \emph{generalize across users}. 

Generalizing robot caregiving across users requires a representation of user functional ability.
For example, consider a robot that is assisting a care recipient with dressing and is deciding where to hold the sleeve opening of a garment.
To choose an appropriate location, the robot must consider both \emph{task success}---whether the user can comfortably reach the sleeve opening---and \emph{agency in action}~\cite{haggard2017sense}--- the user's voluntary actions towards their goals.
Both of these considerations require a detailed and personalized representation of the user.
Designing a new policy for each new user will not scale, and a one-size-fits-all policy will neither succeed in all cases nor empower individual users.

In this work, we consider \emph{functional range of motion (fROM)} of the upper extremities as a central representation for robot caregiving that plays a crucial role in multiple ADLs~\cite{10.5014/ajot.2016.015487} and is strongly correlated with care recipients' general functional ability~\cite{doi:10.1161/STROKEAHA.108.536763}.
A user's fROM is the set of joint positions that they can physically reach without assistance.
Prior work in caregiving robotics, especially in the context of assisted dressing, has collected users' ranges of motion using motion capture~\cite{10.1109/iros.2017.8206206, 10.1109/tro.2019.2904461, 10.1109/access.2020.2978207}.
The requirement of an expensive motion capture system and the need for user data collection present barriers to scaling caregiving robots.

To overcome these barriers, our key insight is that caregiving robots can leverage readily available \emph{functional assessments} from occupational therapy to \emph{predict personalized fROM representations} for care recipients.
These assessments, such as the Action Research Arm Test (ARAT)~\cite{mcdonnell2008action} and the Fugl-Meyer Assessment (FMA)~\cite{Fugl-Meyer} test, are widely used in clinical practice to evaluate upper-extremity function~\cite{castro2014occupational,kane2015functional,murphy2021implementation,jordan2022fast,chen2022predictive}. 

We propose \textbf{GRACE}, a method for learning a generalized predictor of user fROM from functional assessment scores \rv{(Fig. \ref{fig:teaser})}.
We first train a neural encoder that transforms functional assessment scores into a latent representation of user functional ability.
We then use one-class SVM~\cite{scholkopf1999support,Keyvanian2023LearningRJ} to convert sparse fROM data into a dense classification dataset.
From this dataset and the learned functional embeddings, we train a single generalized neural model that classifies whether joint positions are within a user's fROM.
Caregiving robots can use this learned model to predict fROM for new users given only their functional assessment scores.

To train our model, we contribute \textbf{\datasetname}, the first open-source dataset that pairs fROM with functional scores for users with diverse mobility.
We collect data from 11 participants who each emulate 4 different \rv{mobility conditions} using resistance bands, guided by \rv{an occupational therapist}.
Ground-truth reachability labels are collected using motion capture.
Importantly, after our model is trained, motion capture is no longer needed to predict fROM for new users.
Functional scores for each participant in each condition are collected by an OT following standard clinical practice.

We evaluate GRACE with three sets of experiments.
First, we test the generalization capabilities of GRACE in predicting personalized fROM for held-out users and held-out conditions.
Second, we validate the utility of GRACE for robot caregiving with four simulated tasks inspired by handover, rehab, dressing, and bathing.
Finally, we conduct a real-world user study with a real robot and find that GRACE leads to a flexible trade-off between task success and agency in action. 
In summary, our main contributions are as follows:
\begin{itemize}
    \item We show generalized robot caregiving by predicting personalized fROM from functional assessment scores.
    \item We propose GRACE, a generalized neural model that encodes functional assessment scores into a latent user representation and predicts fROM for new users.
    \item We present \datasetname, the first open-source dataset of functional assessment scores and corresponding motion capture data from participants emulating clinically common mobility limitations.
    \item We validate our approach through simulation and real-\rv{robot} experiments, demonstrating its ability to balance task success and user agency in action.
\end{itemize}

\section{Related Work}

\begin{figure*}[!t]
    \centering
    \includegraphics[width=1.0\textwidth]{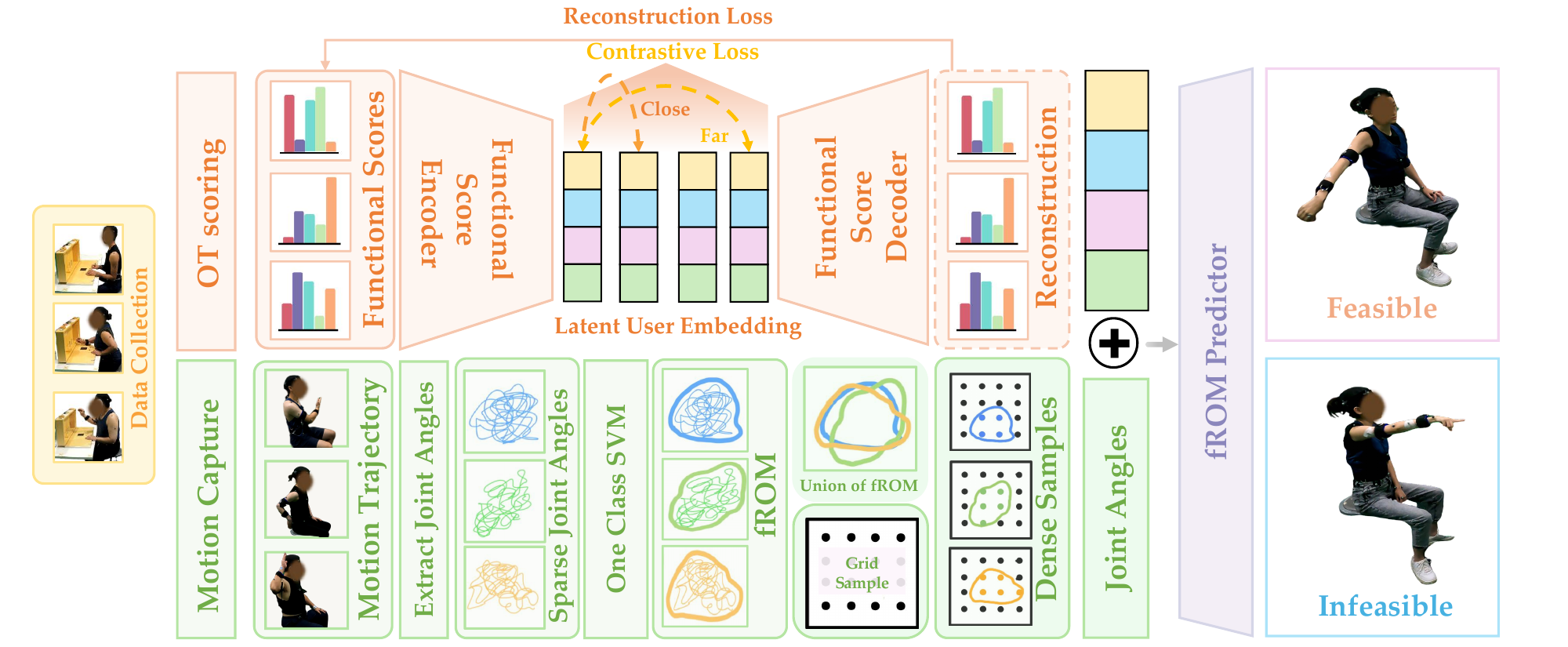}
    \vspace{-0.65cm}
    \caption{Method overview. Using functional assessments scored by occupational therapists (OT) and ground-truth motion capture data, we train a neural model to predict fROM for new users. See text for details.}
    \label{fig:method}
    \vspace{-0.67cm}
\end{figure*}

\subsection{Functional Range of Motion Modeling}

The importance of fROM for general functional ability is well understood in occupational therapy~\cite{doi:10.1161/STROKEAHA.108.536763}.
Activities of daily living such as dressing, grooming, bathing, and feeding, as well as general independence in mobility are greatly affected by and often correlated with fROM~\cite{10.5014/ajot.2016.015487,yamamoto2020relationship}.
Prediction of fROM from clinical functional assessment scores is therefore common in clinical practice.

Modeling fROM has also been considered in the human-robot interaction literature.
For example, Bestick et al. \cite{bestick2015personalized} fit personalized kinematic models for human arms using motion capture data.
Haering et al. \cite{haering2014measurement} take a similar approach while focusing on detailed modeling of the 3 degrees of freedom in the shoulder, again using motion capture.
Other work~\cite{duprey2017kinematic} considers soft-tissue modeling of the human arm as an orthogonal consideration to the personalization and generalization challenges that we address in this work.

Most relevant are prior works that consider fROM modeling for users with mobility limitations.
In particular, we take inspiration from Keyvanian et al. \cite{Keyvanian2023LearningRJ}, who use motion capture to collect fROM data for users with emulated mobility limitations using resistance bands.
This previous work does not consider learning a model that \emph{generalizes across users}.

\subsection{Functional Assessments}
Our central insight in this work is that robot caregiving can be generalized across users by predicting fROM from functional assessment scores.
The connection between functional assessment scores and fROM is well understood in occupational therapy~\cite{OT1,fisher1993improving,manee2020cognitive}.
Functional assessments are also widely used in various fields such as psychology, rehabilitation, and geriatrics~\cite{FA1,FA2,rogers2016functional,wales2012functional} to evaluate an individual's capabilities and limitations in performing everyday tasks and activities. 

Previous work in rehabilitation robotics has considered leveraging functional assessments in combination with machine learning~\cite{ma2020deep,bailey2021machine,campagnini2022machine}.
These works typically consider machine learning as a tool through which functional assessment can be improved.
In contrast, we learn a model that makes predictions \emph{from} functional assessment scores.
To the best of our knowledge, our work is the first to train a machine learning model that predicts fROM from functional assessment scores.

\subsection{Personalization in Robot Caregiving}

The demand for personalized robotic caregiving is multi-dimensional, encompassing both the user's functional abilities and preferences~\cite{madan2022sparcs}. While both are key to improving user satisfaction~\cite{canal2021preferences, jenamani2024feel}, we focus on personalization based on user functional abilities in this paper.

Prior work in robot-assisted dressing directly conditions on user functional abilities to make personalized adjustments based on real-time user movements and mobility limitations~\cite{klee2015personalized, personalization3, personalization4}. For example, Klee et al.~\cite{klee2015personalized} enabled robots to adjust to users' movements and limitations during the dressing process. Zhang et al.~\cite{personalization3} model users in latent spaces to provide personalized assistance for individuals with upper-body impairments. Gao et al.~\cite{personalization4} plan robot motions based on real-time upper-body posture and user models. These methods personalize assistance by actively collecting range of motion data, typically using vision modules during task execution. Beyond dressing assistance, He et al.~\cite{GeneralizationAssistive} propose learning the latent representation of human behavior by training in simulation, and adapting to the preferences and needs of the novel users during task execution at test time. 
In contrast to these works, we propose learning a mapping from functional assessments to range of motion, allowing us to personalize assistance for unseen users based solely on readily available functional assessment scores. This approach eliminates the need for extensive data collection during task execution and enables rapid personalization without burdening the user.

\section{Problem Formulation}

For a caregiving robot to provide personalized assistance while considering a user's agency, it must be able to predict whether a candidate joint configuration falls within the user's \rv{fROM}. In this work, we focus on four degrees of freedom (DOF) in the right arm: shoulder rotation, abduction/adduction, flexion/extension, and elbow flexion. Formally, a joint configuration is a vector of joint angles $\jointstate \in [0, 2\pi)^4$ \rv{obtained from motion capture data}, and we wish to determine the fROM $\from_{\user}$ for each user $\user$, where 
$$\from_{\user} = \{ \jointstate : \jointstate \text{ is reachable for user } \user \}.$$

\vspace{-0.1cm}
When predicting fROM for a user, we assume the availability of their functional assessment scores. In this work, we use ARAT~\cite{mcdonnell2008action} and FMA~\cite{Fugl-Meyer}, which are standard in occupational therapy and can be easily collected in-situ by a licensed therapist. In contrast, directly measuring fROM typically requires a lab setting with expensive motion capture systems~\cite{10.1109/iros.2017.8206206, 10.1109/tro.2019.2904461, 10.1109/access.2020.2978207}. We combine the functional scores for user $u$ into a vector $\functionalscores_{\user} \in \mathbb{R}^n$.
Formally, we wish to learn the mapping $f:\functionalscores_{\user} \mapsto \from_{\user}$, that is, predict a user's fROM from their functional assessment scores.

We consider data-driven methods that can learn this mapping for new users after training on users with known functional scores and fROM.
We assume access to training data of the form $\{(\functionalscores_{\user}, c_{\user}, \from_{\user}) : \user \in \text{training\ users} \}$, where $c_{\user}$ is a discrete mobility condition category (see Section~\ref{subsec:emulating-data}).
We evaluate performance on a dataset of unseen users with unknown and potentially unseen condition categories using normalized Matthews correlation coefficient (nMCC)~\cite{chicco2020advantages}.

\section{A Generalized Model for Personalized fROM}

In this section, we detail GRACE, our novel data-driven method for predicting personalized fROM from functional assessment scores \rv{(Fig. \ref{fig:method})}. First, we reformulate fROM prediction as a binary classification problem (Section \ref{subsec:reformulation_classification}). Next, we learn a compact latent representation that captures a user’s functional assessment results (Section \ref{subsec:latent}). Finally, we train a neural network to predict the feasibility of a given joint configuration conditioned on the latent representation (Section \ref{subsec:neural_network}).

\subsection{Reformulating fROM prediction as Binary Classification}
\label{subsec:reformulation_classification}

To predict the personalized fROM set $\from_{\user}$ for a user $\user$, we reformulate the problem as a binary classification task. The goal is to determine whether a given joint configuration $\jointstate$ is feasible for a specific user, conditioned on their functional assessment scores $\functionalscores_{\user}$.

We begin by defining a bounding volume in the joint space that covers all possible joint configurations observed across participants and conditions. The minimum and maximum values for each joint dimension are computed across all users and conditions, yielding a hyper-rectangular bounding volume, denoted as $\from_{\text{bound}}$. However, the actual feasible fROM for each user is only a subset of this volume. Naively sampling within $\from_{\text{bound}}$ would result in a disproportionate number of negative samples.
To ensure a balance between feasible and infeasible configurations, we construct a mesh grid with 40 samples per joint dimension, spanning the entire $\from_{\text{bound}}$.
\rv{To reduce the high-dimensional joint space, w}e then refine this set by taking the union of all joint configurations that are feasible across all users and conditions, resulting in a reduced, feasible set, denoted as $\from_{\text{union}}$.
Subsequently, for each user, we use $\from_{\user}$ to classify the mesh grid points in $\from_{\text{union}}$ into positive and negative examples.

For each $\jointstate \in \from_{\text{union}}$, we concatenate the user's functional scores $\functionalscores_{\user}$ with the joints $\jointstate$ to create the classification input:
\[
\classificationinput_u = [\functionalscores_{\user}; \jointstate].
\]
The reachability of $\jointstate$ assigns a binary class label $\classificationoutput_u$, indicating whether the joint configuration is feasible for the user $\user$. We aggregate all $(\classificationinput_u, \classificationoutput_u)$ pairs across users to create the classification dataset.

\subsection{Learning a Latent Representation for User Functionality}
\label{subsec:latent}

We extract a compact latent representation to encapsulate a user's functional assessment results using an autoencoder model. The input is \rv{all} normalized functional assessment scores $\functionalscores_{\user} \in \mathbb{R}^n$ for user $\user$, which are embedded into a latent vector $\latent_{\user} \in \mathbb{R}^k$ by the encoder $\mathcal{E}$:
\[
\latent_{\user} = \mathcal{E}(\functionalscores_{\user}).
\]
The latent vector $\latent_{\user}$ represents the user's capabilities and is then passed through the decoder $\mathcal{D}$ to reconstruct the original functional scores:
\[
\hat{\functionalscores}_{\user} = \mathcal{D}(\latent_{\user}).
\]
The encoder and decoder are two-layer feed-forward neural networks, each with a hidden layer size of 16 and a latent vector dimension $k = 4$.
The training objective is to minimize reconstruction error using mean squared error (MSE) loss.
Additionally, a contrastive loss $\mathcal{L}_{\text{contrastive}}$ is applied to shape the latent space, enforcing proximity between embeddings with similar mobility and separation between distinct ones:
\begin{equation*}
\begin{split}
    \mathcal{L}_{\text{contrastive}} = &\frac{1}{|P|} \sum_{(i,j) \in P} \Big[ \mathbb{I}_{\{c_i = c_j\}} \cdot d(\mathbf{z}_i, \mathbf{z}_j)^2 \\
    &+ \mathbb{I}_{\{c_i \neq c_j\}} \cdot \max(0, m - d(\mathbf{z}_i, \mathbf{z}_j))^2 \Big]
\end{split}.
\end{equation*}
where $d(\cdot, \cdot)$ is $L_2$ distance, $m$ is a margin parameter, and $c_i$ and $c_j$ represent the discrete mobility condition categories for users $i$ and $j$ respectively.

\subsection{Predicting Joint Feasibility with a Neural Model}

\label{subsec:neural_network}

After training the functional score encoder, we train a neural network to predict the feasibility of a given joint configuration, conditioned on the latent representation of the user's functional assessment. The model takes as input the concatenation of the latent user embedding $\mathbf{z}_u \in \mathbb{R}^k$ and the joint configuration $\jointstate \in [0, 2\pi)^4$, providing a combined representation $\mathbf{w}_u \in \mathbb{R}^{k+4}$:
\[
\mathbf{w}_u = [\mathbf{z}_u; \jointstate].
\]
This combined vector $\mathbf{w}_u$ serves as the input to a three-layer feed-forward neural network $\mathcal{F}(\mathbf{w}_u; \phi)$ with learnable parameters $\phi$. The output of the network is a scalar $\hat{y}_u \in \{0, 1\}$, which predicts whether the joint configuration $\jointstate$ is feasible for the user $u$:
\[
\hat{y}_u = \mathcal{F}(\mathbf{w}_u; \phi).
\]
The neural network is trained by minimizing a binary cross-entropy loss.
The network consists of three fully connected layers with hidden size 16. We optimize the parameters $\phi$ using Adam~\cite{kingma2014adam} with a learning rate of $5 \times 10^{-4}$, a batch size of 4096, and train the model for 10 epochs.

\section{A New Dataset for fROM Prediction}

\begin{figure*}[!t]
    \centering
    \includegraphics[width=1.0\textwidth, height=0.23\textheight, trim=60mm 95mm 62mm 77mm, clip]{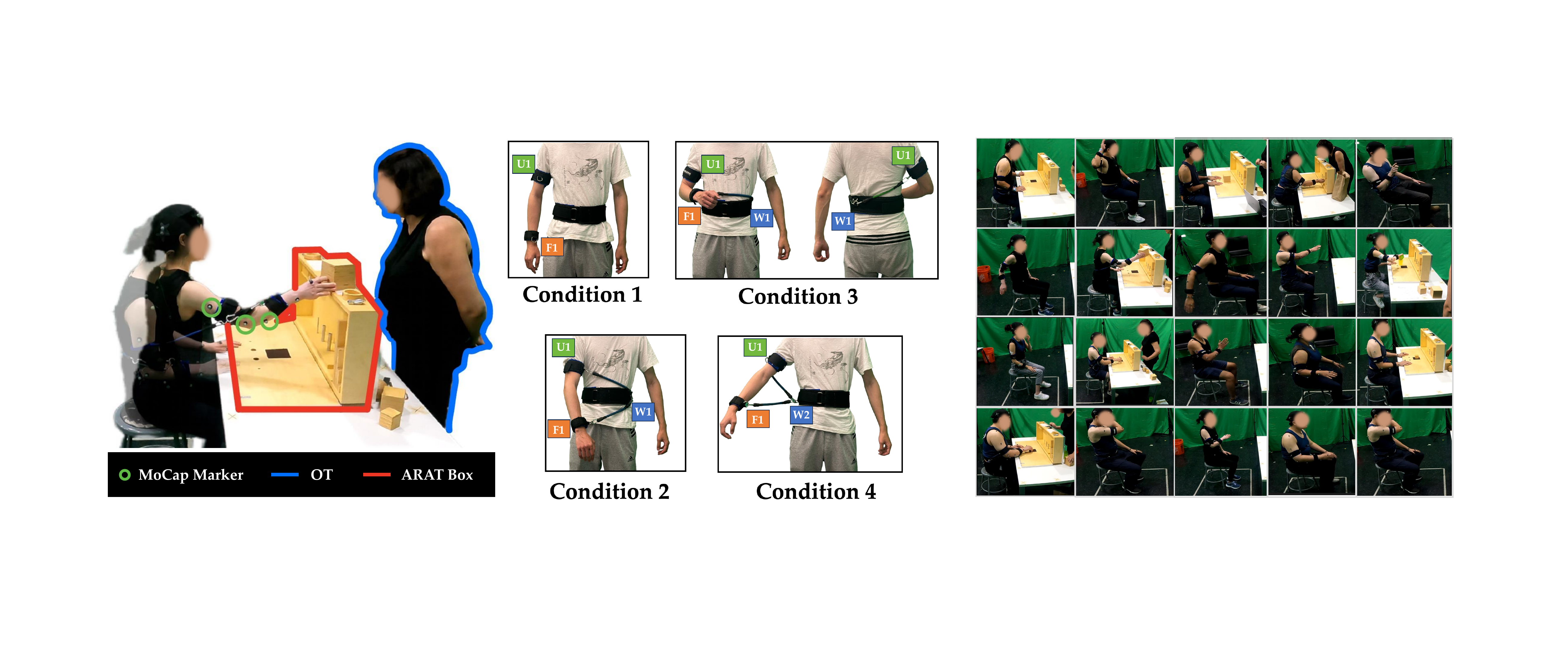}
    \vspace{-0.5cm}
    \caption{\small \textbf{Left}: Collecting data for a subject with emulated mobility limitations using ARAT. \textbf{Center and Right}: We collected functional assessments and fROM data (tracked by motion capture) with 11 subjects, each emulating four conditions. $U_1$, $F_1$, $W_1$ are anchor points on the upper arm, forearm, and waist.}
    \label{fig:data-collection}
    \vspace{-0.65cm}
\end{figure*}

Training GRACE requires a dataset of functional assessment scores and fROM for multiple users with diversity in their mobility.
This paper contributes \datasetname, the first open-source dataset of this kind.
In this section, we describe our procedure for collecting this dataset.

\subsection{Emulating Mobility Limitations with Resistance Bands}
\label{subsec:emulating-data}

\rv{Following common practice in occupational therapy education~\cite{10.1111/1440-1630.12726, 10.1111/1440-1630.12372} and prior HRI work~\cite{Keyvanian2023LearningRJ}, and with guidance from a licensed occupational therapist, we designed our emulation setup using exercise bands of variable resistance (Fig. \ref{fig:data-collection}) to represent clinically relevant mobility limitations. While we acknowledge that real patient data offers the most definitive validation, our occupational therapy collaborator confirmed that measurements taken in these emulated conditions closely resemble the functional limitations of care recipients, providing confidence in the ecological validity of our approach.}

We collected data from 11 participants, each emulating 4 different conditions of mobility, for a total of 43 ``users'' in the dataset (one participant-condition was removed due to technical error). \rv{The participants included 6 East Asian, 1 White, and 4 South Asian individuals; 3 females and 8 males; heights ranging from 160 cm to 185 cm; aged between 20 and 30 years; and varying in exercise routines and body shapes.}

We secured a belt around each participant's waist (resting on the pelvic bone), serving as the anchor point for attaching resistance bands to the right arm. Specifically, 2 anchor points on the belt were secured at the 
pelvic level, $W_1$, located at the left iliac crest, and $W_2$, located at the right iliac crest. Two additional armbands were anchor points on the right arm. The $F_1$ armband was secured at the forearm 1 inch from the radius styloid process of the wrist, and the $U_1$ armband was secured at the midpoint of the humerus.

We vary the anchor points and the resistance level of the exercise bands ($b_{easy}$, $b_{hard}$) to induce mobility constraints that restrict motion across different degrees of freedom of the right arm. 
We selected configurations of the restraints to emulate mobility limitations that closely resemble clinical conditions. For each participant, these configurations created targeted multi-joint limitations, but the severity level of the limitations varied among participants. The conditions are: 
\begin{itemize}
    \item \textbf{Condition 1}: No resistance bands are applied to the participant. This represents no mobility limitation. 
    \item \textbf{Condition 2}: $F_1$ is connected to $W_1$ with the stronger resistance band ($b_{hard}$) in front of the participant’s body. $U_1$ is connected to $W_1$ with $b_{hard}$ in the same direction. This configuration limited shoulder flexion and abduction, elbow flexion, and forearm supination. This mobility limitation pattern is common in stroke survivors (right hemiplegia)~\cite{alt2015kinematic,costa2020intergame}, Spinal Cord Injury C4-C5 \cite{ko2022kinematics}, and severe osteoarthritis in older adults \cite{hartnett2023osteoarthritis}, among other medical conditions.  
    \item \textbf{Condition 3}: $F_1$ is connected to $W_1$ using $b_{hard}$ in front of the participant’s body, while $U_1$ is connected to $W_1$ with the lighter resistance band ($b_{easy}$) behind the participant. This configuration induced shoulder internal rotation, severely restricted shoulder flexion and abduction, limited elbow flexion, and induced forearm pronation. The configuration emulates mobility limitations resulting from severe brain lesions (e.g. stroke, traumatic brain injury)~\cite{mathieu2024upper}, neurodevelopmental disorders (e.g., cerebral palsy)~\cite{rutz2018management},  or neurodegenerative disease conditions (e.g., amyotrophic lateral sclerosis)~\cite{ivy2014upper}.     
    \item \textbf{Condition 4}: $F_1$ is connected to $W_2$ using $b_{hard}$ positioned laterally on the participant’s body, while $U_1$ is also connected to $W_2$ with $b_{hard}$ at the side of the participant’s body. This configuration restricted participants' shoulder full flexion, full abduction, and forearm full supination. This motor pattern is common among individuals with neurodegenerative diseases (e.g., multiple sclerosis)~\cite{bertoni2015unilateral} and orthopedic conditions (e.g., shoulder rotator cuff calcification tendinitis)~\cite{guido2024clinical}.
\end{itemize}
Condition orders were randomized between participants.
For each condition, we repeated the same procedure for collecting functional assessment scores and ground-truth fROM data, which we describe next.

\subsection{Collecting Functional Assessment Scores}
\label{subsec:functional-assessment-scores}

 Functional assessment scores were collected by an experienced licensed occupational therapist following standard clinical practice.
 Another occupational therapist, blind to the purpose of the study, independently scored the recorded clinical assessment videos. Inter-rater reliability was strong across all participants and tasks (Cohen's Kappa 0.9 and above). 
  We selected scores from two functional assessments that are most commonly used across clinical conditions to evaluate upper extremities functional motor performance~\cite{santisteban2016upper, demers2017activity}: ARAT~\cite{mcdonnell2008action} and FMA~\cite{Fugl-Meyer}.

\textbf{Action Research Arm Test (ARAT)} ARAT has four sub-tests: grasp, grip, pinch, and gross movement.
 Given our focus on mobility limitations of the shoulder and elbow joints (inclusive of forearm pronation/supination), we carried out 3 sub-tests.
In the grasp sub-test, participants were instructed to grasp six different objects of varying sizes, shapes, and weights, and move them from table height to the top of the ARAT box (Fig.~\ref{fig:data-collection}). The grip sub-test involved tasks such as pouring water and picking up objects to place them in specific locations inside the ARAT box. In the gross movement sub-test, participants were asked to perform movements such as touching the back of their head, the top of their head, and their mouth. The ARAT items are scored based on observation of movement performance using a 4-level scale.

\textbf{Fugl-Meyer Assessment (FMA)}  The Fugl-Meyer Assessment has 5 sub-tests; in this study, we utilize only the upper extremities sub-tests (FMA-UE), which involve diagonal movement patterns, complex movement patterns, and coordination/speed in movement patterns of the right arm.
The occupational therapist evaluated mobility limitations and task completion following the FMA scoring system.

\subsection{User Study: Collecting Ground-Truth fROM}

We collected ground-truth fROM data using motion capture with 10 OptiTrak cameras. Each subject is fitted with 27 reflective markers on their upper body, arms, and head, as shown in Fig.~\ref{fig:data-collection}.  To minimize inaccuracies in joint angle reconstruction caused by slippage, we attached the markers directly to the participant's skin rather than using a traditional motion capture suit.

The data collection protocol for fROM involved two phases, designed to maximize coverage of the joint configuration space while ensuring repeatability. In phase one, participants replicated guided motions demonstrated by an occupational therapist, ensuring consistent and maximal exploration of joint limits. 
In phase two, participants performed self-directed movements for 30 seconds, with instructions to reach specific spatial landmarks within the scene. This combination of guided and free-motion phases enabled a comprehensive capture of both controlled and naturalistic movement patterns.

\subsection{Extracting Joint Angles from Motion Capture}
We utilize the bone segment poses generated by Motive's human skeleton model to compute the right arm joint angles. Following the International Society of Biomechanics (ISB) standard~\cite{WU2005981} and prior work~\cite{Keyvanian2023LearningRJ}, we derive joint angles through a series of transformations defined as follows.

\textbf{Relative Rotation Matrix} Let $\mathbf{R}_i$ and $\mathbf{R}_j$ be the rotation matrices for segments $i$ and $j$ respectively. The relative rotation matrix $\mathbf{R}_{i,j}$ representing the orientation of segment $j$ relative to segment $i$, is computed as $\mathbf{R}_{i,j} = \mathbf{R}_i^\top \mathbf{R}_j$.

\textbf{Shoulder Joint Angles} For the shoulder joint, the relative rotation matrix is decomposed into a sequence of Euler angles $\jointstate_{\text{shoulder}} = (\theta_{\text{plane}}, \theta_{\text{elev}}, \theta_{\text{rot}})$, corresponding to the plane of elevation, elevation in plane, and shoulder rotation, respectively. This representation follows the ISB-recommended order of rotations for the glenohumeral joint (GHJ)~\cite{WU2005981}.

\textbf{Elbow Joint Angle} 
To determine the elbow joint angle, we extract the primary direction vectors from the upper arm and forearm segments, denoted as $\mathbf{v}_{\text{upper}}$ and $\mathbf{v}_{\text{forearm}}$. We then use their normal vector to find the plane that contains  $\mathbf{v}_{\text{upper}}$ and $\mathbf{v}_{\text{forearm}}$, then project each vector onto the plane and normalize to obtain $\mathbf{u}_{\text{upper}}$ and $\mathbf{u}_{\text{forearm}}$. The relative angle $\theta_{\text{elbow}}$ between the upper arm and forearm is then computed as $\theta_{\text{elbow}} = \cos^{-1}\left( {\mathbf{u}_{\text{upper}} \cdot \mathbf{u}_{\text{forearm}}} \right)$.

\subsection{Completing fROM with One-Class SVM}
\label{subsec:one-class-svm}

The fROM data collected for a given user $\hat{\from}_{\user}$ is a finite subset of $\from_{\user}$, the infinite fROM for that user.
Following previous work~\cite{Keyvanian2023LearningRJ}, we \emph{complete} the fROM by fitting a one-class SVM model~\cite{scholkopf1999support} to $\hat{\from}_{\user}$.
In all experiments, we use an RBF kernel ($\gamma=0.0003, \nu=0.01$) for the SVM.
We then use the decision boundary of the SVM to define $\from_{\user}$.

\section{Experiments}

\begin{figure}[!t]
    \centering
    \includegraphics[width=0.40\textwidth, height=0.22\textheight]{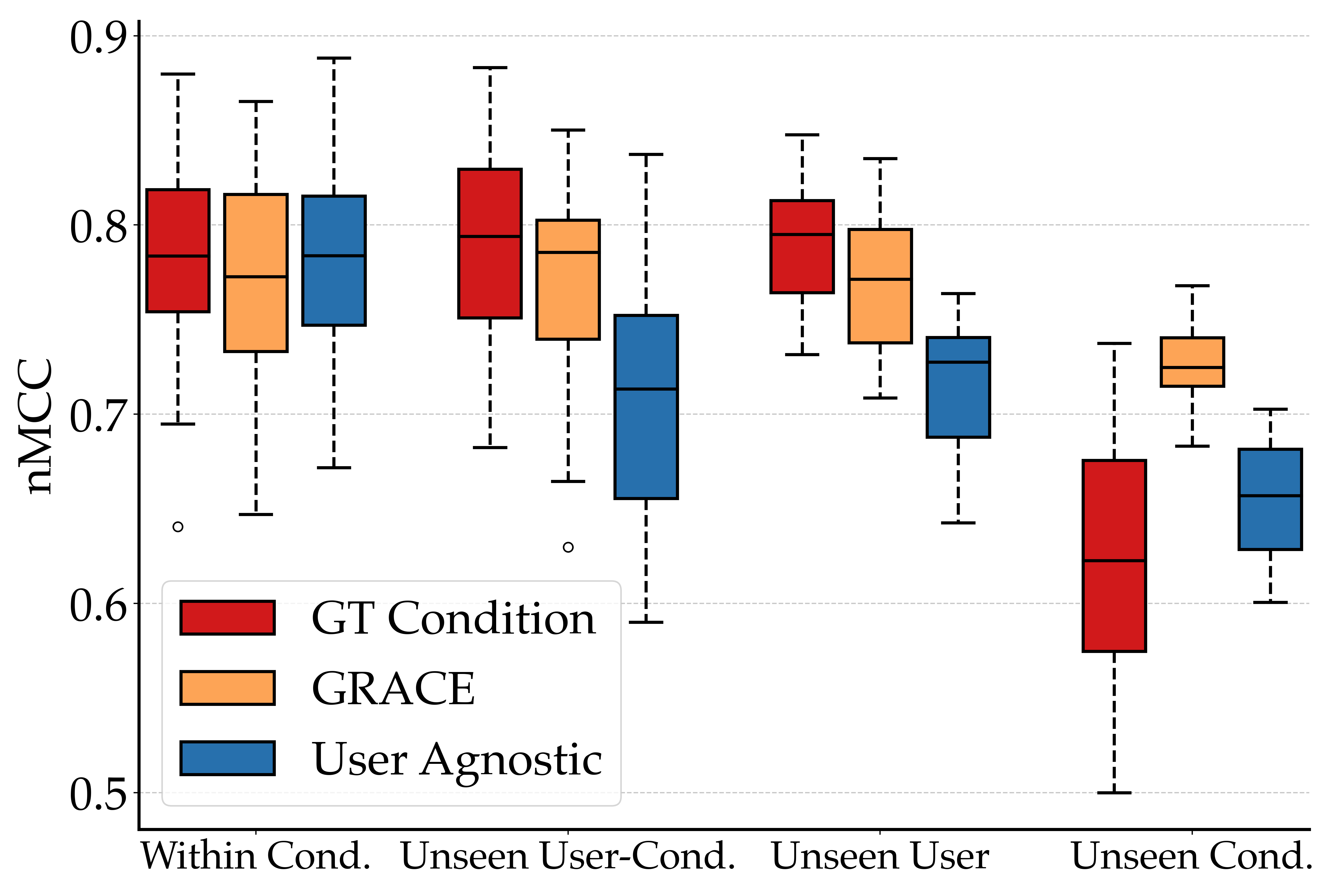}
    \vspace{-0.3cm}
    \caption{\small Evaluation compared with two baselines in four settings shows that GRACE generalizes to unseen users and conditions.}
    \label{fig:exp_prediction}
    \vspace{-2em}
\end{figure}

We are interested in the extent to which GRACE can predict accurate, personalized, and useful fROM from functional assessment scores.
We perform three sets of experiments: a direct empirical analysis, a suite of simulation-based evaluations, and a user study with a real robot arm.

\subsection{Analysis of User Functionality Model}
\label{subsec:exp-a}

We first directly evaluate the ability of our model to predict fROM for new users.
Throughout the three sets of experiments, we use the normalized Matthews Correlation Coefficient (nMCC) as a measure of classification performance.
nMCC is 1.0 for perfect performance and 0.5 in expectation for random predictions, even in the presence of class imbalance.
We compare our method against two baselines:
\begin{enumerate}
    \item \textbf{Ground Truth Condition}: uses ground-truth condition ID instead of learned functional assessment embeddings.
    \item \textbf{User Agnostic}: omits the functional score encoder leading to a non-personalized model.
\end{enumerate}

\noindent In this set of experiments, we test three hypotheses by varying the training and evaluation procedure. \rv{Statistical significance is assessed using the t-test.}

\vspace{4pt}

\emph{\underline{Hypothesis 1}: Leveraging functional assessment scores improves fROM predictions for unseen users.}
We evaluate this hypothesis under two experimental scenarios. 
\rv{In the first scenario, we run leave-one-out cross-validation with the 43 users (Fig. \ref{fig:exp_prediction}).}
\rv{Our method significantly outperforms User Agnostic in nMCC ($p < 0.001$) and is not significantly different from Ground Truth Condition ($p = 0.0869$).}
These results suggest that our method is able to leverage functional scores to outperform a non-personalized model and match a model that has access to ground-truth user condition information.

Recall that our data collection process has each of the 11 participants emulate 4 conditions.
In a second analysis, we select one hold-out \emph{participant} from the 11, train the model on all conditions for the remaining 10 participants, and evaluate on all 4 conditions for the hold-out user.
We perform cross-validation over 5 random seeds.
Results are shown in Fig. \ref{fig:exp_prediction}.
Our model again significantly outperforms the User Agnostic baseline ($p < 0.001$), further indicating that incorporating functional assessment scores yields more accurate fROM predictions for previously unseen users.
In this scenario, our model exhibits slightly lower performance relative to Ground Truth Condition ($p = 0.0012$).

\emph{\underline{Takeaway 1:}}
Our model predicts personalized fROM using functional scores for unseen users \rv{in our dataset}.

\vspace{4pt}

\emph{\underline{Hypothesis 2}: Leveraging functional assessment scores improves personalized fROM predictions for unseen conditions.}
We hold out one of the four conditions during training and evaluate the model on all users in the unseen condition.
We perform cross-validation for each hold-out condition with five random seeds.
Results are shown in Fig.~\ref{fig:exp_prediction}.
Our method significantly outperforms both the Ground Truth Condition ($p < 0.001$) and User Agnostic baseline ($p < 0.001$).

\emph{\underline{Takeaway 2:}} The combination of functional score assessments along with our learned fROM model can generalize to unseen conditions \rv{in our dataset}.

\vspace{4pt}

\emph{\underline{Hypothesis 3}: Our model captures differences in fROM for users with the same underlying condition.}
We train on 10 users with the same condition and evaluate on the held-out user.
We cross-validate and run 5 random seeds.
Statistical tests show no significant difference between our method with the Ground Truth Condition baseline ($p = 0.3296$) and the User Agnostic baseline ($p = 0.3627$). 
Consequently, this hypothesis is nullified: we find no evidence to suggest that our method can distinguish between users with the same condition. 

\emph{\underline{Takeaway 3:}} \rv{While the functional assessments we use in this work are reliable measures of motor function~\cite{10.1191/0269215505cr832oa, 10.1177/1747493017711813}, they may not capture all relevant factors affecting fROM; incorporating additional assessments could improve performance~\cite{nguyen2023correlation}.}

\subsection{Simulated Robot Experiments}

In this set of experiments, we evaluate the extent to which GRACE can be used for personalized and generalized robot caregiving.
We run simulation experiments in four environments inspired by common caregiving scenarios.
In each environment, we define a \emph{generalization scenario}, a \emph{success} criteria, and an \emph{agency in action} metric (higher is better).
We describe the environments briefly here and refer to supplementary materials on our website for details.
\begin{enumerate}
    \item \textbf{Handover}: A user sitting in a wheelchair asks the robot to hand over an object in the environment. The robot selects a 3D handover position.
    \item \textbf{Rehab}: A robot guides a user in a wheelchair through an arm stretching exercise where the robot selects target joint positions and the user attempts to reach them.
    \item \textbf{Dressing}: A robot selects a position at which to hold the arm-hole of a garment. The user attempts to reach that position and then extends their arm through the sleeve.
    \item \textbf{Bathing}: A robot is performing assisted bed bathing. The user's arm must be repositioned. The robot decides whether to ask the user to independently move their arm, or to move their arm for them.
\end{enumerate}

\begin{figure*}[!t]
    \centering
    \includegraphics[width=1.0\textwidth]{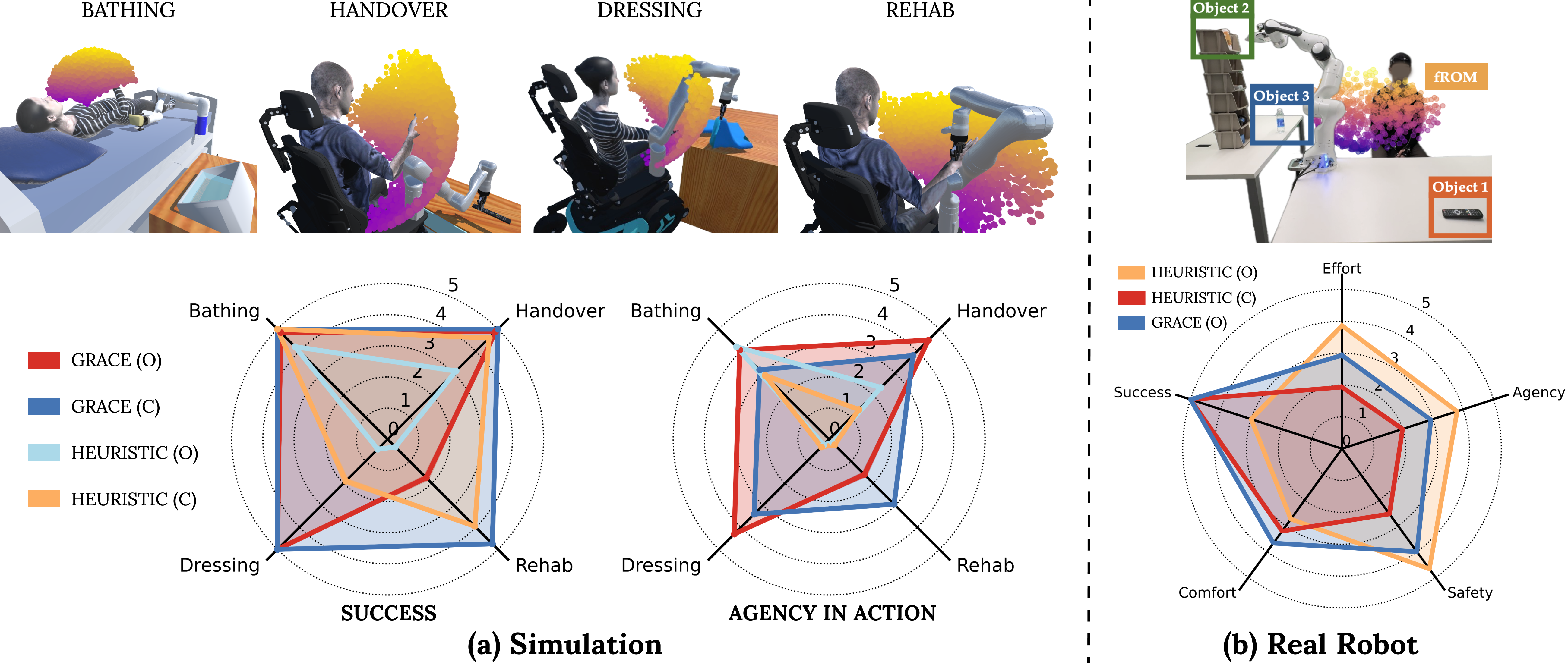}
    \vspace{-0.7cm}
    \caption{\small \textbf{Top Left}: Visualizations of the four environments in our simulation experiments. \textbf{Bottom Left}: Simulation evaluations show that GRACE achieves better performance balancing task success and agency in action compared to heuristic baselines. \textbf{Right}: User study results show that GRACE improves user's sense of agency, physical effort, and safety while providing effective assistance.}
    \label{fig:sim}
    \vspace{-0.67cm}
\end{figure*}

All agency results are normalized with respect to a ground-truth method that has access to the fROM for the held-out user.
We use PyBullet~\cite{coumans2015bullet} for forward kinematics and RCareWorld~\cite{ye2022rcare} for visualization (Fig.~\ref{fig:sim}).
For each environment, we sample 100 tasks and evaluate on 5 held-out users.
We compare the following approaches:
\begin{enumerate}
    \item \textbf{GRACE (Optimistic)}: Use the fROM predicted by our model to maximize agency subject to task success.
    \item \textbf{GRACE (Conservative)}: Same, but predict fROM conservatively: classify reachable if $P(\hat{\classificationoutput}_\user \ge 0.95)$.
    \item \textbf{Heuristic (Optimistic)}: Assume a spherical fROM centered at the resting position with a large radius (30 cm).
    \item \textbf{Heuristic (Conservative)}: Same, but with a smaller radius (10 cm).
\end{enumerate}
Using this setup, we test the following hypothesis:

\emph{\underline{Hypothesis 4}: Personalized fROM prediction can improve user agency and success rates in robot caregiving tasks.}

Simulation results are presented in Fig.~\ref{fig:sim}.
We first find that GRACE consistently outperforms the heuristic baselines.
Furthermore, \rv{we observe that optimism increases agency but reduces success. C}omparing the optimistic and conservative variations of GRACE, we see the method's ability to flexibly balance task success and agency in action.
These results demonstrate that by incorporating user-specific functional data, GRACE tailors assistance to individual needs, thereby empowering users to take a more active role in the caregiving process.

\emph{\underline{Takeaway 4}:} Personalized fROM prediction enables a trade-off between success rate and agency in robot caregiving.

\subsection{Real-Robot User Study}

In this final set of experiments, we conduct a user study with a real robot.
The setup is shown in Fig.~\ref{fig:sim}.
We consider a handover task involving three objects---a TV remote, a prescription bottle, and a water bottle---based on a list of household objects prioritized for robot retrieval by people with ALS~\cite{5209484}.
The robot is a Franka Emika Panda 7-DOF arm with polymetis controllers~\cite{lin2021polymetis} running on an Ubuntu 20.04 desktop.
We recruit 5 participants from the original 11 and have each of them again emulate mobility limitations with resistance bands (Conditions 2, 3, and 4; see Section~\ref{subsec:emulating-data}).

For each participant and for each of the three objects, we use GRACE (Optimistic), Heuristic (Optimistic), and Heuristic (Conservative) to select handover positions. For GRACE, \rv{we choose the optimistic variant as it best represents the overall performance in simulation}, and use a model trained on all participants except for the current one. \rv{Additional comparisons with GT Condition and User Agnostic baselines (see Section~\ref{subsec:exp-a}) are in the supplementary material on our website.}
In each trial, the robot picks the object and brings it to the handover position proposed by the model.
The user then attempts to reach the object and answers a series of questions about the interaction.
We conduct this user study in two parts, where the order of trials varies to address two different hypotheses.
The first part tests the following hypothesis:

\emph{\underline{Hypothesis 5}: Personalized fROM prediction improves task success rates, sense of agency, physical effort, and safety in a real robot-human handover task.}

We first evaluate all three methods and all three objects for a participant emulating \emph{a single condition}.
The order of methods and objects is counterbalanced  to mitigate ordering effects.
We consider the following metrics:
\begin{enumerate}
    \item \textbf{Task Success}: whether the participant is able to grasp the object after handover (true or false).
    \item \textbf{Sense of Agency}: the participant's perception of the extent to which they are actively using their physical abilities (5 point Likert scale).
    \item \textbf{Physical Effort}: the participant's self-reported physical exertion (5 point Likert scale).
    \item \textbf{Safety}: the proximity of the robot gripper to the user's hand during object handover (5 point Likert scale).
    \item \textbf{Comfort} the user's self-reported comfort during object handover (5 point Likert scale).
\end{enumerate}

Results are shown in Fig.~\ref{fig:sim}.
We see that GRACE significantly outperforms the Heuristic (Conservative) baseline in terms of sense of agency ($p = 0.0054$), physical effort ($p = 0.0080$), and safety ($p < 0.001$).
GRACE also significantly outperforms the Heuristic (Optimistic) in terms of task success ($p = 0.0018$), \rv{with slightly lower but comparable safety ($p = 0.0069$).}
Other differences were not statistically significant.
These findings are consistent with our simulation results and underscore the flexible trade-off between task success and other metrics that our method enables.

\emph{\underline{Takeaway 5}:} GRACE improves task success, sense of agency, physical effort, and safety for real users.

\noindent The second part of the study tests the hypothesis:

\emph{\underline{Hypothesis 6}: Users perceive GRACE's fROM prediction to be personalized.}

To test this hypothesis, we vary emulated conditions for each user.
The robot performs one handover with the water bottle for each condition to evaluate each approach.
We ask participants to rank the approaches in terms of perceived personalization.
We found that all five users consistently ranked GRACE the highest in terms of personalization.

\emph{\underline{Takeaway 6}: GRACE's personalization aligns with real users' perception of personalization.}

Altogether, this real-world user study demonstrates that GRACE significantly enhances the user's sense of agency and encourages more active physical participation while also improving safety during robot interactions---all without compromising task success.
Additionally, users consistently rated GRACE as more adaptive and responsive to their individual abilities compared to the heuristic baseline, highlighting the method's ability to predict personalized fROM from functional assessment scores, making it more responsive and empowering for users with mobility limitations.

\section{Discussion}

In this work, we considered the challenge of scaling robot caregiving by \emph{generalizing across users}.
Our key insights were that fROM serves as a general-purpose representation for personalizing robot policies and can be predicted from readily available functional assessment scores.
We introduced \datasetname, the first open-source dataset with paired functional assessment scores and fROM data, and used it to train GRACE, a neural model that predicts fROM from functional assessment scores.
In three sets of experiments, we found that GRACE consistently outperforms baselines and enables personalized assistance in multiple robot caregiving scenarios.

There remain many open challenges for personalizing robot caregiving.
For fROM prediction, we found that GRACE was unable to significantly distinguish between users within the same mobility condition based on their functional scores. \rv{GRACE can be extended to integrate additional assessment scores that capture an even wider range of functional limitations. While these functional scores reliably reflect fROM over time for stable conditions, we encourage future research to consider transient or gradual changes in user functionality, where scores are frequently reassessed.}

For personalizing robot caregiving more broadly, this work represents a small step---much remains to be done.
Our efforts here demonstrate the viability of generalizing robot caregiving through learned user models and suggest a path forward to answer the call for personalization in assistive robotics~\cite{jenamani2024feel,bedaf2018multi,madan2022sparcs,nanavati2023physically}.

\newpage
\newpage

\bibliographystyle{IEEEtran} 
\balance
\bibliography{references}

\end{document}